\documentclass[runningheads]{llncs}
\usepackage{graphicx}
\usepackage{subfigure}
\usepackage{booktabs} 
\usepackage{amsmath}
\usepackage{bbm}
\usepackage{array}
\usepackage{wrapfig}
\usepackage{amssymb}
\usepackage{fontawesome}
\usepackage{multirow}
\usepackage{float}
\begin{document}
%
\title{Calibrated Diverse Ensemble Entropy Minimization for Robust Test-Time Adaptation in Prostate Cancer Detection}

\author{Mahdi Gilany\inst{1}\inst{\star}, Mohamed Harmanani\inst{1}, Paul Wilson\inst{1}, Minh Nguyen Nhat To\inst{2}, Amoon Jamzad\inst{1}, Fahimeh Fooladgar\inst{2}, Brian Wodlinger\inst{3}, Purang Abolmaesumi\inst{2}, Parvin Mousavi\inst{1}}
\institute{Corresponding Author: mahdi.gilany@queensu.ca \\
\inst{1}School of Computing, Queen’s University, Kingston, Canada \\ \inst{2}Department of Electrical and Computer Engineering, University
of British Columbia, Vancouver, Canada \\
\inst{3}Exact Imaging, Markham, Canada}
\maketitle
\begin{abstract}
High resolution micro-ultrasound has demonstrated promise in real-time prostate cancer detection, with deep learning becoming a prominent tool for learning complex tissue properties reflected on ultrasound. However, a significant roadblock to real-world deployment remains, which prior works often overlook: model performance suffers when applied to data from different clinical centers due to variations in data distribution. This distribution shift significantly impacts the model's robustness, posing major challenge to clinical deployment. Domain adaptation and specifically its test-time adaption (TTA) variant offer a promising solution to address this challenge. 
In a setting designed to reflect real-world conditions, we compare existing methods to state-of-the-art TTA approaches adopted for cancer detection, demonstrating the lack of robustness to distribution shifts in the former.
We then propose Diverse Ensemble Entropy Minimization (DEnEM), questioning the effectiveness of current TTA methods on ultrasound data. We show that these methods, although outperforming baselines, are suboptimal due to relying on neural networks output probabilities, which could be uncalibrated, or relying on data augmentation, which is not straightforward to define on ultrasound data. Our results show a significant improvement of $5\%$ to $7\%$ in AUROC over the existing methods and $3\%$ to $5\%$ over TTA methods, demonstrating the advantage of DEnEM in addressing distribution shift. 
\keywords{Ultrasound Imaging \and Prostate Cancer \and Computer-aided Diagnosis \and Distribution Shift Robustness \and Test-time Adaptation.}
\end{abstract}

\section{Introduction}
Prostate cancer (PCa) diagnosis remains a pivotal challenge, where early and accurate detection can significantly influence treatment outcomes. The standard diagnostic approach is transrectal ultrasound-guided biopsy (TRUS), which \textit{systematically} samples biopsy cores from uniform locations across the prostate. However, this method cannot specifically target suspicious areas due to similar acoustic tissue properties and low resolution, often resulting in missed diagnoses or unnecessary interventions.
Alternatively, advanced ultrasound modalities aim to improve tissue characterization and offer targeted diagnostic strategies. 
In particular, high-frequency micro-ultrasound (micro-US) has recently emerged for visualization of prostate tissue at significantly higher resolutions than conventional imaging~\cite{ghai2016assessing}, facilitating the identification of potential cancer lesions. Using a qualitative approach, micro-US has demonstrated comparable PCa detection to multi-parametric MRI targeted biopsy~\cite{abouassaly2020impact,cotter2023comparing}, with the distinct advantage of operating in real-time. However, quantitative approaches interpreting this modality is in early-stage research with few studies~\cite{gilany2022towards,rohrbach2018high} proposed for user-independent and objective PCa detection. 
Recent studies have utilized deep learning to analyze micro-US images for identifying PCa~\cite{gilany2023trusformer,wilson2023self}. 
While deep learning can effectively extract complex tissue features from noisy ultrasound images, its adaptability to any data distribution can undermine model \textit{robustness} during deployment when data distributions shift. 
Such shifts significantly impact model performance, as the model may learn biases in the training distribution and form spurious correlations between input images and tissue types~\cite{koh2021wilds,sagawa2019distributionally}. 
This study argues that, despite recent successes, the performance and effectiveness of existing micro-US PCa detection studies are overestimated.
These studies typically assume similar data distributions between training and deployment scenarios, which is unrealistic in clinical settings where shifts in data distribution are likely to occur due to differences in patient populations, operator techniques, and imaging devices.

To improve PCa detection robustness and mitigate distribution shifts, we explore the promising field of domain adaptation, where techniques are developed to adjust trained models to perform well on a target domain with a shift in distribution~\cite{kouw2019review}. Specifically, we investigate a more challenging variant, \textit{test-time adaptation} (TTA), which employs unsupervised objectives to fine-tune trained models on a single test sample during inference~\cite{liang2023comprehensive}. This approach eliminates the need to access the entire test dataset and accounts for variability in data distributions of test patients or biopsy cores. 
In practice, several challenges hinder the direct application of state-of-the-art (SOTA) TTA methods in PCa detection. Recent studies indicate that prior methods may fail in real-world scenarios with multiple types of distribution shifts~\cite{niu2023towards}, and their effectiveness varies depending on the shift type~\cite{zhao2023pitfalls}. Moreover, 
the most common and promising TTA methods utilize either self-supervised learning (SSL) or entropy loss in their core~\cite{sun2020test,wang2020tent}.
However, choice of SSL data augmentation for ultrasound is not straightforward, and current entropy-based methods rely on neural networks generated probabilities (thus, entropy), known to be biased and uncalibrated~\cite{guo2017calibration}. This issue is exacerbated by distribution shifts and ultrasound artifacts and noisy labels~\cite{ovadia2019can}.
An ideal TTA method should be tailored for micro-US shifts, produce calibrated entropy, and not rely on augmentations. To this end, we propose DEnEM, a novel TTA method for addressing clinical center distribution shifts in micro-US PCa detection. DEnEM utilizes an ensemble of neural networks diversified through mutual information minimization to produce calibrated marginal entropy without the need for data augmentations. Our key contributions are:
\begin{enumerate}
    \item We investigate the existing SOTA PCa detection methods in a real-world setting, revealing the impact of distribution shift on their performance.
    \item We propose leveraging SOTA TTA methods, for the first time, to address clinical center distribution shifts in PCa detection. We demonstrate their effectiveness in significantly improving over existing methods while also examining their limitations on micro-US data.
    \item Inspired by TTA methods, we propose DEnEM, a novel approach tailored for micro-US that addresses entropy calibration and the reliance on data augmentations. We demonstrate that DEnEM significantly outperforms both baselines and SOTA TTA methods across several evaluation metrics. 
\end{enumerate}

\section{Related Works}
\noindent\textbf{Micro-US PCa detection:} 
Recent studies have employed quantitative approaches for micro-US PCa detection~\cite{gilany2022towards,rohrbach2018high,shao2020improving,wilson2024toward}. Rohrbach et al.\cite{rohrbach2018high} introduced the first quantitative method using SVMs on manually extracted ultrasound spectral features, which is less effective than deep learning and requires a reference phantom, limiting its usability. Gilany et al.\cite{gilany2022towards} demonstrated the success of CNNs in automatically extracting features from RF images to detect PCa. More recent methods~\cite{gilany2023trusformer,wilson2023self} addressed labeled data scarcity and weak labeling, improving detection performance. However, these studies often neglect clinical integration and robust deployment across various clinical centers, leading to overestimated and unrealistic performance due to shifting data distributions.

\noindent\textbf{Test-time Adaptation:} 
TTA is a challenging form of unsupervised domain adaptation, requiring the adaptation of a model trained on a source domain to a shifted target domain during inference, without accessing the \textit{source domain} to preserve data privacy, and without the \textit{entire target domain}, enabling immediate application post-deployment \cite{liang2023comprehensive}.
The core of TTA lies in defining a proxy objective to adapt the model to test samples unsupervised. These objectives include entropy minimization \cite{wang2020tent,zhang2022memo}, self-supervised learning (SSL) \cite{bartler2022mt3,sun2020test}, and feature alignment \cite{zhao2023pitfalls}.
``TENT"~\cite{wang2020tent} fine-tunes the model, at inference, by minimizing the entropy on test prediction probabilities $p$: $H(p) = -\sum_cp^c~\text{log}~p^c$. ``MEMO"~\cite{zhang2022memo} extends TENT by minimizing the entropy of \textit{marginal} probability distribution calculated across a set of augmentations. While entropy minimization is a theoretically well established TTA objective~\cite{goyal2022test}, neural networks often produce biased and uncalibrated entropy~\cite{guo2017calibration}. Moreover, ``TTT"~\cite{sun2020test} and ``MT3"~\cite{bartler2022mt3} use SSL proxy objectives. 
These methods modify the neural architecture by attaching a self-supervision head to the feature extractor network. During inference, this head and SSL objective adapt the feature extractor to the test distribution. However, these approaches have two drawbacks: they alter the neural architecture, potentially degrading performance in some tasks, and they only fine-tune the feature extractor, leaving the classification head unadapted. The latter is particularly impactful on distribution shift~\cite{kirichenko2022last}.

\section{Materials and Method}
\subsection{Data}
Our study employs a dataset collected from five clinical centers, with consented data from 693 patients who underwent systematic TRUS-guided prostate biopsy as part of a clinical trial (NCT02079025 clinicaltrials.gov). The procedure leverages high-frequency micro-US technology, typically capturing 10-12 biopsy cores per patient. Prior to firing the biopsy gun, a single RF image of the prostate with $28~mm$ depth and $46~mm$ width is captured. These images, paired with corresponding histopathology reports of the cores
constitute our dataset. 
The reports include a binary label indicating the presence of cancer, an estimate of the cancer length or ``involvement of cancer", and the aggressiveness of cancer as determined by the Gleason Score \cite{MICHALSKI20161038}. Overall, our dataset includes $6607$ biopsy cores with $5727$ of those being benign. More details of the data distribution among centres are provided in the supplementary material.  

The needle trace region in the RF image, where histopathology labels are assigned, is first identified~\cite{gilany2022towards,gilany2023trusformer,shao2020improving}. From this region, we perform patch extraction by capturing overlapping regions of interest (ROIs) that correspond to a tissue area of $5~mm$ by $5~mm$, using $1~mm$ by $1~mm$ strides (see Figure \ref{main_figure} (a)). An RF patch is considered to be within the needle region if there is at least a 60\% overlap, a threshold set to compensate for potential inaccuracies in identifying the needle region~\cite{wilson2023self}. RF patches are resized from their original dimensions of $1780\times55$ to $256\times256$ to align with our deep network architectures.

\subsection{Method}
\begin{figure}[tb]
\centering
\includegraphics[width=1.\linewidth]{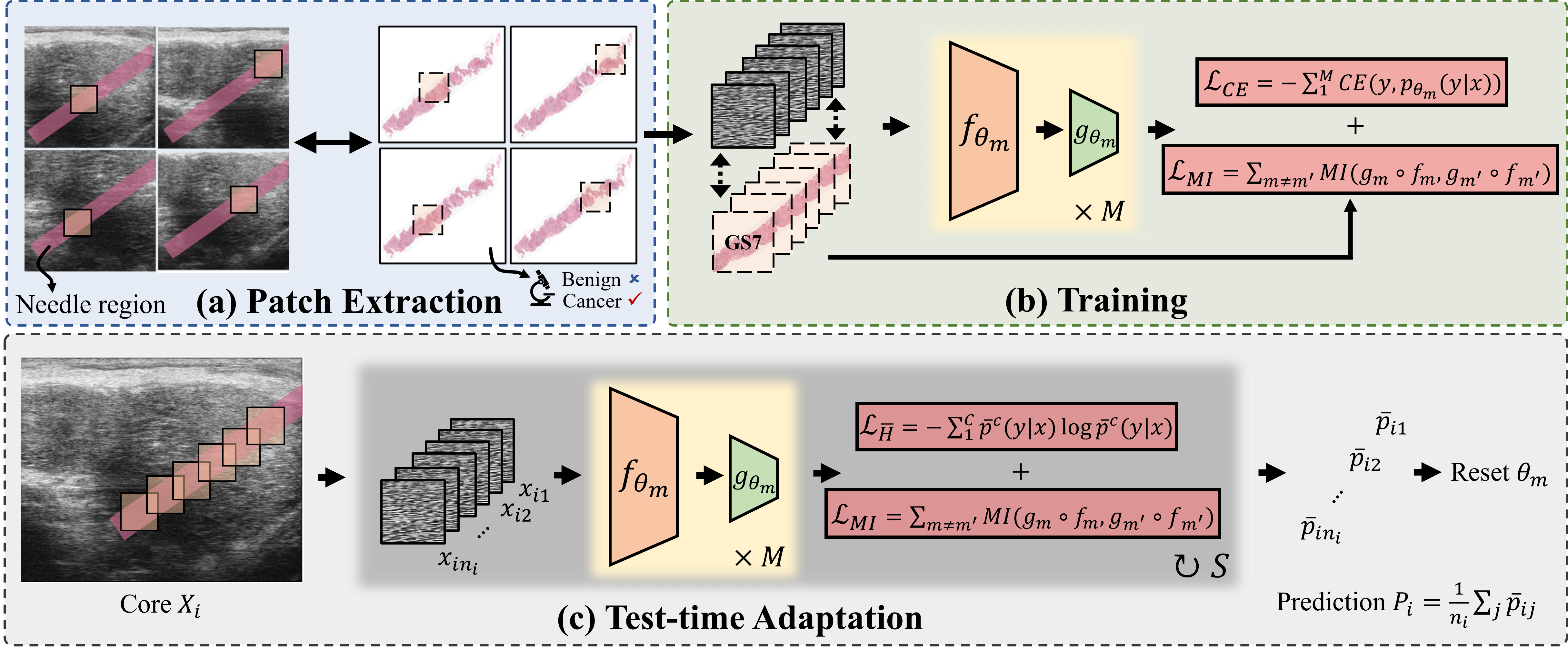}%
\caption{
Overview of DEnEM method. (a) RF patches extraction from needle region. (b) Deep ensemble training with cross entropy and mutual information losses. (c) Model adaptation at inference to each core with marginal entropy loss before the prediction.
}
\label{main_figure}
\end{figure}

Figure~\ref{main_figure} shows an overview of our proposed approach. This approach trains a deep ensemble~\cite{lakshminarayanan2017simple} of neural networks to minimize both cross-entropy loss and mutual information loss. During inference, marginal entropy is utilized to adapt the deep ensemble to the distribution of patches from a biopsy core in test set.      

\noindent\textbf{Deep Ensemble:} Our model borrows ResNet10~\cite{he2016deep} image encoder from prior PCa detection literature~\cite{gilany2022towards,wilson2023self}. Deep ensemble is  a collection of $M$ encoders and classifiers with different random weight initialization. Unlike single network models, deep ensemble not only provides a highly calibrated entropy, but also this calibration remains less affected by distribution shift~\cite{ovadia2019can}.

\noindent\textbf{Mutual Information Loss:} Randomly initialized neural networks in deep ensemble may still learn similar encoders and classifiers, providing biased and uncalibrated entropy. To overcome this, we propose ensuring statistical independence of predictions across the networks. Inspired by "DivDis"~\cite{lee2022diversify}, we minimize the mutual information between the predictions of each network pair. The mutual information loss ($\mathcal{L}_{MI}$) and the total training loss ($\mathcal{L}$) are defined as follows:
\begin{align}
    \mathcal{L}_{MI}(g_m\circ{f_m},g_{m'}\circ{f_{m'}})&=\mathcal{D_{KL}}(p_{\theta_m,\theta_{m'}}(y_m,y_{m'})||p_{\theta_m}(y_m)\otimes{p_{\theta_{m'}}(y_{m'})})\\
        \mathcal{L}=\mathcal{L}_{CE}&+\lambda\sum_{m\neq{m'}}\mathcal{L}_{MI}(g_m\circ{f_m},g_{m'}\circ{f_{m'}})
\end{align}
where $p_{\theta_m}(y|x)$ is the output probability distribution of a data sample from the $m$'th model with parameters $\theta_m$. $\mathcal{D_{KL}}$ refers to $\mathcal{KL}$ divergence between joint probability distribution of models $m$ and $m'$, and is calculated empirically~\cite{lee2022diversify}. Note that unlike ``DivDis", our mutual info. loss is calculated on the same training/adaptation data.  $\mathcal{L}_{CE}$ represents the sum of cross-entropy loss for all networks, and $\lambda=10$ controls the importance of mutual info. loss. 

\noindent\textbf{Marginal Entropy Loss:} Entropy minimization~\cite{wang2020tent} and marginal entropy minimization~\cite{zhang2022memo} are promising proxy objectives for test-time robustness and adaptation to distribution shift with theoretically well established literature~\cite{goyal2022test}. However, neural networks may not provide a well calibrated entropy, significantly undermining the effectiveness of these methods. Additionally, data augmentations may not be feasible or properly studied for a specific data type like ultrasound RF images, proposing new challenges to the problem. Therefore, we propose to find marginal entropy across deep ensemble networks instead of across various augmentations~\cite{zhang2022memo}. We first find marginal distribution by averaging $M$ output probabilities: $\overline{p}(y|x)=\mathbb{E}_M[p_{\theta_m}(y|x)]=\frac{1}{M}\sum_{m=1}^Mp_{\theta_m}(y|x)$. Then, the marginal entropy loss is calculated as: $\mathcal{L}_{\overline{H}} =-\sum_{c=1}^C\overline{p}^c(y|x)\times\text{log}~\overline{p}^c(y|x)$.

\noindent\textbf{Test-time Adaptation:} To adapt to test distribution, we fine-tune the ensemble networks by minimizing both unsupervised objectives of marginal entropy ($\mathcal{L}_{\overline{H}}$) and mutual info. ($\mathcal{L}_{MI}$) losses on RF patches of each test biopsy core $X_i = \{x_{i1}, x_{i2}, ..., x_{i{n_i}}\}$, separately. After $S$ iteration of adaptation, with learning rate $lr_{adapt}$, the parameters of ensemble networks, i.e. $\theta_m$ are recovered to the values after initial training. This is equivalent to episodic adaptation on each biopsy core, and we leave online adaptation for future work due to the challenging hyperparameter tuning~\cite{zhao2023pitfalls}. Lastly, we follow SOTA TTA methods~\cite{wu2018group} and substitute all batch norm layers \cite{ioffe2015batch} with batch-agnostic group norm layers \cite{wu2018group}. More discussions on this effective approach are presented in the next sections.




\section{Experiments}
We organize our experiments in following parts: 
(i) \textit{SOTA baseline comparison}. 
We compare our model's performance with prior SOTA methods for PCa detection in micro-US. This includes patch classification with single or with an ensemble of ResNet10~\cite{gilany2022towards}, patch classification with SSL pre-training on unlabeled data~\cite{wilson2023self}, and core classification with TRUSFormer~\cite{gilany2023trusformer} which collects patch extracted features with a transformer aggregator.  
(ii) \textit{SOTA TTA comparison.}
Additionally, we compare our model against SOTA TTA methods, including TENT~\cite{wang2020tent} with ResNet10 backbone, MEMO~\cite{zhang2022memo} with marginal entropy calculated across various augmentations proposed in \cite{wilson2023self}, TTT~\cite{sun2020test} with BYOL~\cite{grill2020bootstrap} as the SSL objective, and its extended meta learning version MT3~\cite{bartler2022mt3}. We also include ``SAR" which filters unreliable samples based on entropy and minimizes a smoothed entropy loss~\cite{niu2023towards}.
(iii) \textit{Ablation studies}. We perform two ablation studies and a qualitative analysis. First, we evaluate the contribution of each of our model's component to the performance. Second, we follow TTA methods~\cite{niu2023towards} and study the effect of batch-agnostic norm layers. In particular, we replace ResNet10 batch norm~\cite{ioffe2015batch} layers with group norm~\cite{wu2018group} and compare.

\noindent\textbf{Evaluation Strategy:} 
We evaluate all methods using a leave-one-center-out (LOCO) strategy, holding all data from one clinical center for testing in each experiment. This approach reflects real-world deployment. The remaining four centers are divided into training and validation with patient-wise stratification. To ensure balanced representation, the dataset from each center is first divided into training and validation sets and then combined.
The performance is measured at core-level by averaging output probabilities of RF patches inside a core. Following previous works~\cite{rohrbach2018high,gilany2022towards}, cores with cancer involvement $\leq40\%$ are removed from test for more stable evaluation and AUROC and balanced-accuracy (sensitivity and specificity average) are reported. We additionally include AUROC-All where all test set cores, regardless of cancer involvement, are included.

\noindent\textbf{Implementation Details:}
We re-implement all baseline models to evaluate them in a realistic LOCO setting with data distribution shift. We use PyTorch 2.1, with training and validation of our method taking approximately 4.6 hours on a single NVIDIA A40 GPU. Inference on a biopsy core, including adaptation, takes around $250~ms$. We manually tune hyperparameters based on validation. Adam optimizer~\cite{kingma2014adam} with learning rate of $1e-4$ and a scheduler with cosine annealing and a linear warm-up were used. For adaptation during inference, we employed SGD with the best learning rate selected from $lr_{adapt}=\{1e-1, 1e-2, 1e-3\}$ and number of iterations from $S=\{1, 5\}$. Additional details are provided in the supplementary document.

\section{Results and Discussion}
\begin{table}[tb]
\caption{PCa detection performance for baselines, TTAs, and our proposed method. 
}\label{tab:maintable}%
\centering
\begin{tabular}{@{}lc@{\hspace{.5cm}}c@{\hspace{.5cm}}c@{}cc@{}}
\toprule
\bf Method & AUROC & AUROC-All  & Balanced-Acc. \\ 
\midrule
\textbf{Baselines}\\

ResNet10 & $75.2 \pm 7.0$ & $68.3 \pm 6.0$ & $68.0 \pm 4.6$ \\ 

Ensem.~\cite{lakshminarayanan2017simple} & $75.8 \pm6.4$ & $68.1\pm 6.3$&  $68.0 \pm 6.2$\\

SSL + ResNet \cite{wilson2023self} & $75.1 \pm 4.0$ & $67.7 \pm 5.3$ & $68.0 \pm 4.3$\\

TRUSFormer \cite{gilany2023trusformer}  & $75.3\pm 4.9$ & $70.4 \pm 2.6$ & $68.8 \pm 4.0$ \\
\midrule
\textbf{Test-time adaptations}\\
TENT \cite{wang2020tent} & $77.3 \pm 4.2$ & $71.0 \pm 3.6$ & $69.7 \pm 5.4$ \\
MEMO \cite{zhang2022memo} & $77.4 \pm 4.2$ & $70.9 \pm 3.6$ & $69.4 \pm 6.4$   \\ 
TTT \cite{sun2020test} & $77.8 \pm 5.1$ & $71.1 \pm 3.4$ & $66.5 \pm8.8$ \\
MT3 \cite{bartler2022mt3} & $77.7 \pm5.7$ & $70.3 \pm 4.4$ &  $63.3 \pm4.8$  \\
SAR \cite{niu2023towards} & $77.6 \pm 4.4$ & $71.2 \pm3.7$ & $70.7 \pm 5.0$ \\
\midrule 





DEnEM (ours) & $\textbf{80.9} \pm \textbf{4.5}$ & $\textbf{75.0} \pm \textbf{2.9}$ & $\textbf{75.6} \pm \textbf{3.4}$ \\

\bottomrule
\end{tabular}
\end{table}
\noindent\textbf{SOTA Baseline Comparison:}
Table \ref{tab:maintable} summarizes the results of SOTA comparison. Overall, our proposed method (DEnEM) outperforms all baselines by a significant margin in core AUROC, AUROC-All, and Balanced-ACC by around $5\%$ to $7\%$. In particular, our patch classification approach outperforms SOTA core classification TRUSFormer by ${\sim}5\%$ indicating the effectiveness of DEnEM. Future works may leverage the orthogonal benefits of these two methods. Moreover, comparing the baselines shows that the leading method in PCa detection may fall behind relatively simpler approaches in LOCO setting, highlighting the importance of this realistic evaluation. Lastly, note that the high standard deviations are due to averaging the results of \textit{different} test datasets with high variability in performance. In ablation study, we will show the standard deviation for each individual test datasets.



\noindent\textbf{SOTA TTA Comparison:}
Similar to baseline comparisons, our proposed method substantially improves over SOTA TTA methods in all metrics by $3\%$ to $5\%$. Comparing TTA methods show that TENT and MEMO produce similar results, indicating that marginal entropy across various augmentations is not effective. Considering this in conjunction with SSL-pretraining results, we hypothesize that the proposed augmentations for RF images~\cite{wilson2023self} might be ineffective. In conclusion, the performance of DEnEM compared with marginal differences between TTA results justifies the proposed marginal entropy in DEnEM.        

\begin{figure}[tb]
\centering
\includegraphics[width=1.\linewidth]{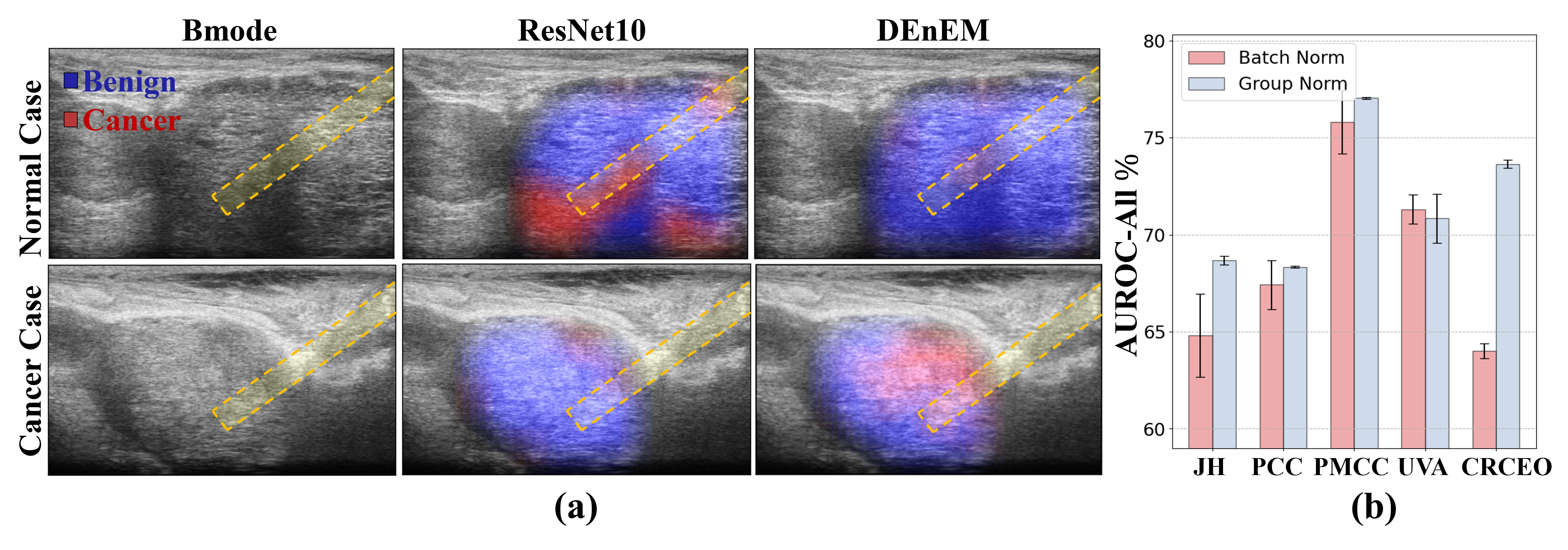}%
\caption{
(a) Heatmap comparison of ResNet10 and DEnEM with cancer (red) vs. benign (blue) areas. Top row: benign core; bottom row: cancerous core (Gleason score 3+4). (b) Baseline ResNet10 Batch norm vs. group norm comparison for different test center. 
}
\label{heatmaps}
\end{figure}

\noindent\textbf{Ablation Studies:}
\begin{table}[tb]
\caption{Ablation study on different components of the proposed method.}\label{tab:ablation}%
\centering
\begin{tabular}{@{}l@{\hspace{0.0cm}}c@{\hspace{.1cm}}c@{\hspace{.1cm}}c@{\hspace{.1cm}}c@{\hspace{.1cm}}c@{}}
\toprule
\bf Method & Group-Norm & Ensem. & Mutual-Info. & Test-Adapt & AUROC-All \\ 
\midrule

ResNet10 & \checkmark & - & - & - & $71.7 \pm 3.7$ \\

Ensem. & \checkmark & \checkmark & - & - & $72.9 \pm 4.1$ \\

Ensem.+$\mathcal{L}_{MI}$ & \checkmark &  \checkmark &  \checkmark & - & $73.8 \pm 3.5$\\

Ensem.+$\mathcal{L}_{\overline{H}}$& \checkmark & \checkmark & - &  \checkmark & $73.1 \pm 4.0$\\

Ensem.+$\mathcal{L}_{MI}$+$\mathcal{L}_{\overline{H}}$ (ours) & \checkmark & \checkmark &  \checkmark &  \checkmark &  $75.0 \pm 2.9$\\

\bottomrule
\end{tabular}
\end{table}
Table \ref{tab:ablation} shows the effect of different components on the performance of our method. Overall, each component increases the performance individually, with $\mathcal{L}_{MI}$ improving by $0.9\%$ and $\mathcal{L}_{\overline{H}}$ improving by $0.2\%$. However, when combined together, the gain is compounded with improvement of $2.1\%$. Unlike the baselines, deep ensemble improves AUROC-All over ResNet10 when group norm is adopted. This expected outcome demonstrates the harmful effect of batch norm on baselines with LOCO setting. Recent TTA methods~\cite{niu2023towards} have also highlighted the potential drawbacks of batch norm under distribution shift as the mean and variance estimation in batch norm layers will be biased. In Figure~\ref{heatmaps} (b), we compare the AUROC-All of ResNet10 using batch norm versus group norm (used in our experiment) with separate bar plots for each center left out for test. This figure further confirms the substantial improvement of replacing batch norm with group norm. Additionally, it reveals a high variability in performance, ranging from $68\%$ to $77\%$ across different centers. We hypothesize that the quantity and quality of data in both training and test sets contribute to this variability, though further research is needed in future works.

\noindent\textbf{Qualitative analysis:}
We qualitatively analyze our model's prediction by using heatmaps, comparing the predictions when no TTA is adopted (e.g. ResNet10 baseline) with DEnEM, as detailed in Figure \ref{heatmaps} (a). These heatmaps are created by running the model across the entire RF image using sliding patches. They showcase two examples of benign (top row) and cancerous (bottom row) cores with red indicating cancer predictions and blue for benign predictions. These maps, overlaid on corresponding B-mode images, not only demonstrate the failure of baseline compared to DEnEM, but also qualitatively show the capability of our model in localizing cancer and offering reliable guidance for targeted biopsy. 




\section{Conclusion}
This study examined the robustness of deep learning models to clinical center distribution shifts in micro-US PCa detection. We showed that existing methods are vulnerable to these shifts, with leading approaches sometimes outperformed by simpler ones. To address this, we adopted TTA, which improved detection but faced limitations due to ultrasound augmentations and biased entropy of neural networks. We proposed DEnEM, which calculates calibrated marginal entropy across a diverse ensemble of networks without requiring data augmentations. DEnEM significantly enhanced performance, outperforming previous TTA methods and demonstrating its potential to improve PCa detection robustness. 

\bibliographystyle{splncs04}
\bibliography{mybibliography}

\begin{thebibliography}{10}
\providecommand{\url}[1]{\texttt{#1}}
\providecommand{\urlprefix}{URL }
\providecommand{\doi}[1]{https://doi.org/#1}

\bibitem{abouassaly2020impact}
Abouassaly, R., Klein, E.A., El-Shefai, A., Stephenson, A.: Impact of using 29 mhz high-resolution micro-ultrasound in real-time targeting of transrectal prostate biopsies: initial experience. World journal of urology  \textbf{38}(5),  1201--1206 (2020)

\bibitem{bartler2022mt3}
Bartler, A., B{\"u}hler, A., Wiewel, F., D{\"o}bler, M., Yang, B.: Mt3: Meta test-time training for self-supervised test-time adaption. In: International Conference on Artificial Intelligence and Statistics. pp. 3080--3090. PMLR (2022)

\bibitem{cotter2023comparing}
Cotter, F., Perera, S., Sathianathen, N., Lawrentschuk, N., Murphy, D., Bolton, D.: Comparing the diagnostic performance of micro-ultrasound-guided biopsy versus multiparametric magnetic resonance imaging-targeted biopsy in the detection of clinically significant prostate cancer: A systematic review and meta-analysis. Soci{\'e}t{\'e} Internationale d'Urologie Journal  \textbf{4}(6),  465--479 (2023)

\bibitem{ghai2016assessing}
Ghai, S., Eure, G., Fradet, V., Hyndman, M.E., McGrath, T., Wodlinger, B., Pavlovich, C.P.: Assessing cancer risk on novel 29 mhz micro-us images of the prostate: creation of the micro-us protocol for prostate risk identification. The Journal of Urology  \textbf{196}(2),  562--569 (2016)

\bibitem{gilany2022towards}
Gilany, M., Wilson, P., Jamzad, A., Fooladgar, F., To, M.N.N., Wodlinger, B., Abolmaesumi, P., Mousavi, P.: Towards confident detection of prostate cancer using high resolution micro-ultrasound. In: International Conference on Medical Image Computing and Computer Assisted Intervention. pp. 411--420. Springer (2022)

\bibitem{gilany2023trusformer}
Gilany, M., Wilson, P., Perera-Ortega, A., Jamzad, A., To, M.N.N., Fooladgar, F., Wodlinger, B., Abolmaesumi, P., Mousavi, P.: Trusformer: improving prostate cancer detection from micro-ultrasound using attention and self-supervision. International Journal of Computer Assisted Radiology and Surgery pp.~1--8 (2023)

\bibitem{goyal2022test}
Goyal, S., Sun, M., Raghunathan, A., Kolter, J.Z.: Test time adaptation via conjugate pseudo-labels. Advances in Neural Information Processing Systems  \textbf{35},  6204--6218 (2022)

\bibitem{grill2020bootstrap}
Grill, J.B., Strub, F., Altch{\'e}, F., Tallec, C., Richemond, P., Buchatskaya, E., Doersch, C., Avila~Pires, B., Guo, Z., Gheshlaghi~Azar, M., et~al.: Bootstrap your own latent-a new approach to self-supervised learning. Advances in Neural Information Processing Systems  \textbf{33},  21271--21284 (2020)

\bibitem{guo2017calibration}
Guo, C., Pleiss, G., Sun, Y., Weinberger, K.Q.: On calibration of modern neural networks. In: International Conference on Machine Learning. pp. 1321--1330 (2017)

\bibitem{he2016deep}
He, K., Zhang, X., Ren, S., Sun, J.: Deep residual learning for image recognition. In: Proceedings of the IEEE Conference on Computer Vision and Pattern Recognition. pp. 770--778 (2016)

\bibitem{ioffe2015batch}
Ioffe, S., Szegedy, C.: Batch normalization: Accelerating deep network training by reducing internal covariate shift. In: International conference on machine learning. pp. 448--456. pmlr (2015)

\bibitem{kingma2014adam}
Kingma, D.P., Ba, J.: Adam: A method for stochastic optimization. arXiv preprint arXiv:1412.6980  (2014)

\bibitem{kirichenko2022last}
Kirichenko, P., Izmailov, P., Wilson, A.G.: Last layer re-training is sufficient for robustness to spurious correlations. arXiv preprint arXiv:2204.02937  (2022)

\bibitem{koh2021wilds}
Koh, P.W., Sagawa, S., Marklund, H., Xie, S.M., Zhang, M., Balsubramani, A., Hu, W., Yasunaga, M., Phillips, R.L., Gao, I., et~al.: Wilds: A benchmark of in-the-wild distribution shifts. In: International Conference on Machine Learning. pp. 5637--5664. PMLR (2021)

\bibitem{kouw2019review}
Kouw, W.M., Loog, M.: A review of domain adaptation without target labels. IEEE transactions on pattern analysis and machine intelligence  \textbf{43}(3),  766--785 (2019)

\bibitem{lakshminarayanan2017simple}
Lakshminarayanan, B., Pritzel, A., Blundell, C.: Simple and scalable predictive uncertainty estimation using deep ensembles. Advances in neural information processing systems  \textbf{30} (2017)

\bibitem{lee2022diversify}
Lee, Y., Yao, H., Finn, C.: Diversify and disambiguate: Out-of-distribution robustness via disagreement. In: The Eleventh International Conference on Learning Representations (2022)

\bibitem{liang2023comprehensive}
Liang, J., He, R., Tan, T.: A comprehensive survey on test-time adaptation under distribution shifts. arXiv preprint arXiv:2303.15361  (2023)

\bibitem{MICHALSKI20161038}
Michalski, J.M., Pisansky, T.M., Lawton, C.A., Potters, L.: Chapter 53 - prostate cancer. In: Gunderson, L.L., Tepper, J.E. (eds.) Clinical Radiation Oncology (Fourth Edition), pp. 1038--1095.e18. Elsevier, Philadelphia, fourth edition edn. (2016)

\bibitem{niu2023towards}
Niu, S., Wu, J., Zhang, Y., Wen, Z., Chen, Y., Zhao, P., Tan, M.: Towards stable test-time adaptation in dynamic wild world. In: The Eleventh International Conference on Learning Representations (2023)

\bibitem{ovadia2019can}
Ovadia, Y., Fertig, E., Ren, J., Nado, Z., Sculley, D., Nowozin, S., Dillon, J., Lakshminarayanan, B., Snoek, J.: Can you trust your model's uncertainty? evaluating predictive uncertainty under dataset shift. Advances in neural information processing systems  \textbf{32} (2019)

\bibitem{rohrbach2018high}
Rohrbach, D., Wodlinger, B., Wen, J., Mamou, J., Feleppa, E.: High-frequency quantitative ultrasound for imaging prostate cancer using a novel micro-ultrasound scanner. Ultrasound in medicine \& biology  \textbf{44}(7),  1341--1354 (2018)

\bibitem{sagawa2019distributionally}
Sagawa, S., Koh, P.W., Hashimoto, T.B., Liang, P.: Distributionally robust neural networks for group shifts: On the importance of regularization for worst-case generalization. arXiv preprint arXiv:1911.08731  (2019)

\bibitem{shao2020improving}
Shao, Y., Wang, J., Wodlinger, B., Salcudean, S.E.: Improving prostate cancer (pca) classification performance by using three-player minimax game to reduce data source heterogeneity. IEEE Transactions on Medical Imaging  \textbf{39}(10),  3148--3158 (2020)

\bibitem{sun2020test}
Sun, Y., Wang, X., Liu, Z., Miller, J., Efros, A., Hardt, M.: Test-time training with self-supervision for generalization under distribution shifts. In: International conference on machine learning. pp. 9229--9248. PMLR (2020)

\bibitem{wang2020tent}
Wang, D., Shelhamer, E., Liu, S., Olshausen, B., Darrell, T.: Tent: Fully test-time adaptation by entropy minimization. arXiv preprint arXiv:2006.10726  (2020)

\bibitem{wilson2023self}
Wilson, P.F., Gilany, M., Jamzad, A., Fooladgar, F., To, M.N.N., Wodlinger, B., Abolmaesumi, P., Mousavi, P.: Self-supervised learning with limited labeled data for prostate cancer detection in high frequency ultrasound. IEEE Transactions on Ultrasonics, Ferroelectrics, and Frequency Control  (2023)

\bibitem{wilson2024toward}
Wilson, P., Harmanani, M., To, M., Gilany, M., Jamzad, A., Fooladgar, F., Wodlinger, B., Abolmaesumi, P., Mousavi, P.: Toward confident prostate cancer detection using ultrasound: a multi-center study. International Journal of Computer Assisted Radiology and Surgery  (2024)

\bibitem{wu2018group}
Wu, Y., He, K.: Group normalization. In: Proceedings of the European conference on computer vision (ECCV). pp. 3--19 (2018)

\bibitem{zhang2022memo}
Zhang, M., Levine, S., Finn, C.: Memo: Test time robustness via adaptation and augmentation. Advances in Neural Information Processing Systems  \textbf{35},  38629--38642 (2022)

\bibitem{zhao2023pitfalls}
Zhao, H., Liu, Y., Alahi, A., Lin, T.: On pitfalls of test-time adaptation. arXiv preprint arXiv:2306.03536  (2023)

\end{thebibliography}

\end{document}


%
\title{Calibrated Diverse Ensemble Entropy Minimization for Robust Test-Time Adaptation in Prostate Cancer Detection}

\author{Mahdi Gilany\inst{1}\inst{\star}, Mohamed Harmanani\inst{1}, Paul Wilson\inst{1}, Minh Nguyen Nhat To\inst{2}, Amoon Jamzad\inst{1}, Fahimeh Fooladgar\inst{2}, Brian Wodlinger\inst{3}, Purang Abolmaesumi\inst{2}, Parvin Mousavi\inst{1}}
\institute{\inst{\star}Corresponding Author: mahdi.gilany@queensu.ca \\
\inst{1}School of Computing, Queen’s University, Kingston, Canada \\ \inst{2}Department of Electrical and Computer Engineering, University
of British Columbia, Vancouver, Canada \\
\inst{3}Exact Imaging, Markham, Canada}

\maketitle              

\section{Appendix}

\subsection{Dataset}

\begin{table}
\centering
\caption{Summary of the dataset. Clinical center names are abbreviated~\cite{shao2020improving}.}
\begin{tabular}{@{}lccccccc@{}}
\toprule
Center & Patients & Cores & Benign & GS7 & GS8 & GS9 & GS10 \\
\midrule
JH      & 60 & 616  & 568  & 32  & 10 & 6 & 0  \\
UVA    & 236 & 2335 & 2018 & 221 & 57 & 28 & 11  \\
PCC    & 171 & 1599 & 1400 & 162 & 23 & 14 & 0 \\
PMCC   & 71  & 588  & 486  & 90  & 12 & 0 & 0 \\
CRCEO  & 155 & 1469 & 1255 & 170 & 32 & 12 & 0 \\
\midrule
Total & 693 & 6607 & 5727  & 675 & 134 & 60 & 11 \\
\bottomrule
\end{tabular}
\end{table}

\subsection{Experiments}
TTA methods were implemented adhering closely to their documented implementation. We modified the TTT method \cite{sun2020test} by substituting its self-supervision task with (Bootstrap Your Own Latent) BYOL \cite{grill2020bootstrap}, inline with MT3. For MEMO, two augmentations proved sufficiently effective. The augmentations for TTT, MT3, and MEMO were random transformations of RF patches through horizontal flip, vertical flip, cropping, and affine transformation proposed in~\cite{wilson2023self}. 
Table below lists some of the hyperparameters used for the experiments. For further details, we will publish our code in near future.


\begin{table}[t]
\centering
\caption{Summary of hyperparameters. We pick the best for the sets.}
\begin{tabular}{@{}lcc@{}}
\toprule
\bf Method & Hyper-parameter & Value \\
\midrule
\multirow{4}{*}{All} & Epochs & 50 \\
& Batch size & $\{32, 128\}$ \\
& Learning rate & $1e-4$ \\
& Scheduler & Cosine annealing w/ linear warmup \\ 
\midrule
\multirow{6}{*}{TTA} & $lr_{adapt}$ & $\{1e-1, 1e-2, 1e-3\}$ \\ 
& $S$ & $\{1, 5, 10\}$ \\
& Norm layer & Group-norm \\
& Num augmentations & 2 \\
& \multirow{2}{*}{Augmentations} & horizontal flip, vertical flip \\
&  & cropping, and affine transformation \\
\midrule
\multirow{2}{*}{DEnEM} & $\lambda$ & $\{0.1, 1, 10\}$ \\
& Num ensembles & 5 \\

\bottomrule
\end{tabular}
\end{table}

\bibliographystyle{splncs04}
\bibliography{mybibliography}